\documentclass{article}

\usepackage[final]{neurips_2025}

\usepackage[utf8]{inputenc} 
\usepackage[T1]{fontenc}    
\usepackage{hyperref}       
\usepackage{url}            
\usepackage{booktabs}       
\usepackage{amsfonts}       
\usepackage{nicefrac}       
\usepackage{microtype}      
\usepackage{xcolor}         
\usepackage{multirow}
\usepackage{amsmath}
\usepackage{graphicx}
\usepackage{caption}
\usepackage{subcaption}
\usepackage{stmaryrd}
\usepackage{listings}
\usepackage{wrapfig}
\usepackage{comment}

\title{QiMeng-NeuComBack: Self-Evolving Translation from IR to Assembly Code}
\author{
  \textbf{Hainan Fang$^{1,2}$, Yuanbo Wen$^{1}$, Jun Bi$^{1}$, Yihan Wang$^{1,2}$, Tonghui He$^{1,2}$, } \\
  \textbf{Yanlin Tang$^{1,2}$, Di Huang$^{1}$, Jiaming Guo$^{1}$, Rui Zhang$^{1}$, Qi Guo$^{1}$, Yunji Chen$^{1, 2}$\thanks{Corresponding Author.} } \\[0.35em]
  $^{1}$State Key Lab of Processors, Institute of Computing Technology, Chinese Academy of  \\
  Sciences, Beijing, China $^{2}$University of Chinese Academy of Sciences, Beijing, China \\[0.35em]
  \texttt{\{fanghainan23s, wenyuanbo, bijun, wangyihan25s, hetonghui23s, } \\
  \texttt{tangyanlin21s, huangdi, guojiaming, zhangrui, guoqi, cyj\}@ict.ac.cn}
}

\begin{document}

\maketitle

\begin{abstract}
Compilers, while essential, are notoriously complex systems that demand prohibitively expensive human expertise to develop and maintain. The recent advancements in Large Language Models (LLMs) offer a compelling new paradigm: Neural Compilation, which could potentially simplify compiler development for new architectures and facilitate the discovery of innovative optimization techniques.
However, several critical obstacles impede its practical adoption. Firstly, a significant lack of dedicated benchmarks and robust evaluation methodologies hinders objective assessment and tracking of progress in the field. Secondly, systematically enhancing the reliability and performance of LLM-generated assembly remains a critical challenge.
Addressing these challenges, this paper introduces NeuComBack, a novel benchmark dataset specifically designed for IR-to-assembly compilation. Leveraging this dataset, we first define a foundational Neural Compilation workflow and conduct a comprehensive evaluation of the capabilities of recent frontier LLMs on Neural Compilation, establishing new performance baselines.
We further propose a self-evolving prompt optimization method that enables LLMs to iteratively evolve their internal prompt strategies by extracting insights from prior self-debugging traces, thereby enhancing their neural compilation capabilities.
Experiments demonstrate that our method significantly improves both the functional correctness and the performance of LLM-generated assembly code.
Compared to baseline prompts, the functional correctness rates improved from 44\% to 64\% on x86\_64 and from 36\% to 58\% on aarch64, respectively.
More significantly, among the 16 correctly generated x86\_64 programs using our method, 14 (87.5\%) surpassed \texttt{clang-O3} performance.
These consistent improvements across diverse architectures (x86\_64 and aarch64) and program distributions (NeuComBack L1 and L2) validate our method's superiority over conventional approaches and its potential for broader adoption in low-level neural compilation.

\end{abstract}

\section{Introduction}
Compilers are indispensable and highly complex software systems, meticulously engineered over decades to translate high-level programming languages into low-level machine code executable by specific hardware architectures \citep{lattner2004llvm}. This process involves multiple stages of analysis, transformation, and optimization, often requiring deep expertise and significant development effort. In recent years, Large Language Models (LLMs) have made significant progress in natural language processing \citep{openai2024gpt4o, liu2024deepseek, yang2024qwen2, grattafiori2024llama} and increasingly excel at code-related tasks such as auto-completion, program synthesis, and bug detection, sparking interest in their broader software engineering applications \citep{hou2024large, wang2024software,rozière2024codellamaopenfoundation,10.1609/aaai.v38i19.30185}.

Given their proficiency in handling complex code-related tasks, a natural and compelling research direction is the exploration of LLMs as core components in compilation (i.e., \textbf{Neural Compilation}).
By directly translating high-level source code or intermediate representations (IR) into low-level assembly code using LLMs, \emph{Neural Compilation} has the potential to augment or even replace traditional compiler stages.
Compared to rule-based compilation, \emph{Neural Compilation} has two appealing benefits. 
Firstly, it can drastically reduce the time and effort needed to build compilers for emerging Instruction Set Architectures (ISAs) \citep{armengol2021learning}. 
Secondly, it can uncover new optimization strategies based on input program semantics due to LLM's ability to process code as lossless text \citep{cummins2023large}.
Thus, \emph{Neural Compilation} has the potential to transform how programming language researchers and hardware architects explore new designs. 

Despite the escalating interest and conceptual feasibility demonstrated by early works \citep{armengol2021learning}, \emph{Neural Compilation} still faces major practical challenges, particularly in ensuring functional correctness and optimizing performance.
First, achieving semantic equivalence comparable to traditional compilers remains a substantial challenge, even for LLMs trained on vast code corpora \citep{cummins2024meta}. 
Second, while LLMs might uncover novel optimization strategies \citep{cummins2023large}, consistently surpassing the optimization levels of mature compilers (e.g., \texttt{clang-O3}) is an even greater hurdle.
Moreover, the absence of systematic benchmarks for evaluating LLMs' \emph{Neural Compilation} capabilities hinders objective progress assessment. Addressing this gap requires (1) a well-defined dedicated benchmark for consistent and measurable criteria, and (2) methods to improve the reliability and performance of LLM-aided neural compilation.

To this end, in this paper, we first introduce \texttt{NeuComBack}, a novel benchmark dataset specifically designed for evaluating IR-to-assembly compilation. Derived from \texttt{ExeBench} \citep{armengol2022exebench} and \texttt{TSVC} \citep{maleki2011evaluation}, \texttt{NeuComBack} provides a diverse set of programs to systematically assess fundamental compilation and optimization capabilities. Note that the benchmark focuses on the critical backend compilation task of translating LLVM IR to hardware-specific assembly code across different ISAs. Using this benchmark, we conduct a systematic evaluation of state-of-the-art LLMs, including DeepSeek-R1~\citep{guo2025deepseek}, establishing crucial performance baselines that were previously unavailable in this emerging research area.
Second, we define a foundational \emph{Neural Compilation} workflow, encompassing generation and optimization steps, which serves as a basis for our experiments and can inform future research in this domain. Building upon this workflow, we propose an automatic prompt learning method for assembly generation that learns from the LLM's self-debugging traces, systematically extracting insights from generation and correction attempts and iteratively evolving internal prompt strategies.

The main contributions of this work are summarized as follows:

\begin{itemize}
    \item We introduce \texttt{NeuComBack}, a novel benchmark dataset tailored for the LLVM IR-to-assembly \emph{Neural Compilation} task.
    Based on it, we conduct a comprehensive evaluation of state-of-the-art LLMs, establishing critical and previously unavailable baselines.

    \item We propose a self-evolving prompt optimization technique that enables LLMs to automatically improve their compilation capability through iterative refinement.
    By analyzing compilation errors and performance optimization trails, our method dynamically adapts prompt strategies, enhancing both the correctness and the performance.
    
    \item We conduct extensive experiments in terms of the functional correctness and the performance of LLM-generated assembly code. Compared to the baseline prompt, correctness increased from 44\% to 64\% (x86\_64) and 36\% to 58\% (aarch64). More importantly, 14 of 16 correct x86\_64 programs (87.5\%) surpassed the performance of \texttt{clang-O3}.
    These results demonstrate consistent gains across diverse architectures and benchmarks, validating our approach's superiority in neural compilation.
    
\end{itemize}

\vspace{8ex}
\section{Related Work}

\subsection{Neural Compilation}

Machine learning has a long history in compiler optimization, traditionally focusing on heuristics for tasks like pass selection and phase ordering \citep{wang2018machine, agakov2006using, trofin2021mlgo}.
A more ambitious direction, \emph{Neural Compilation}, aims for end-to-end assembly code generation. Early efforts in this vein employed neural machine translation techniques.
For instance, \citet{armengol2021learning} pioneered direct C-to-x86 assembly translation using Transformers, and \citep{hosseini2022beyond, armengol2024slade} developed specialized models for decompilation.

The advent of LLMs has significantly advanced neural compilation. 
\citet{cummins2023large} demonstrated that LLMs, trained on millions of  LLVM assembly and corresponding optimal compiler optimizations, could predict and perform the beneficial optimizations.
Based on that, \citet{cummins2024meta} further introduced LLM Compiler, a family of foundation models pretrained on compiler IRs and assembly.
These models facilitate downstream tasks like compiler flag tuning and disassembly.
LLMs are also increasingly applied to other compiler-level tasks such as sophisticated decompilation \citep{tan2024llm4decompile, wong2023refining} and compiler fuzzing \citep{deng2023large, yang2024whitefox}.
\citet{fang2024towards} explored whether advanced reasoning mechanisms, like chain-of-thought \citep{wei2022chain}, could enhance LLM performance in applying a single peephole optimization on AArch64 assembly. Their findings suggest that conventional LLMs, even when fine-tuned, struggle due to a lack of reasoning, and that models augmented with explicit reasoning capabilities (e.g., GPT-o1-preview) can significantly outperform them\citep{wang2023chatgptdefendbelieftruth}. 

Our work builds upon these diverse efforts in leveraging machine learning for compilation.
While prior studies have demonstrated the initial feasibility of neural translation for compilation and showcased the growing power of LLMs in both optimizing and generating low-level code, achieving robust functional correctness and competitive performance simultaneously in \emph{Neural Compilation} remains a primary research objective.
As defined in Section~\ref{sec:NeuralCompilationTask}, our research aims at translating LLVM IR to assembly code, targeting both functional correctness and performance optimization. 

\subsection{Automatic Prompt Learning}

The performance of Large Language Models (LLMs) heavily depends on input prompts, making prompt engineering crucial. While manual techniques like zero-shot, few-shot prompting \citep{radford2019language, brown2020language,wei2024chatiezeroshotinformationextraction,10.5555/3600270.3600412}, and Chain-of-Thought (CoT) \citep{wei2022chain,zhang2022automaticchainthoughtprompting,zhao2024enhancingzeroshotchainofthoughtreasoning,zhou2023threadthoughtunravelingchaotic,wang2023selfconsistencyimproveschainthought} exist, their laborious creation and often suboptimal or inconsistent results have spurred research into Automatic Prompt Optimization (APO).

APO methods algorithmically discover effective prompts through iterative refinement, sometimes using evolutionary approaches or LLMs as optimizers \citep{zhou2022large, pryzant2023automatic, yang2023large, guo12connecting,shum2024automaticpromptaugmentationselection,sun2023autohintautomaticpromptoptimization}.
\citet{gao2025prompt} introduced MAPS for test case generation, focusing on LLM-tailored prompts through diversity, domain knowledge, and failure-driven rule induction. \citet{ye2025prompt} proposed Prochemy for code generation/translation, using execution-driven iterative refinement to produce a fixed, reusable, optimized prompt. These highlight a trend towards automated, task-aware, and model-specific prompt engineering for code.

Our Automatic Prompt Learning (APL) method (Section~\ref{sec:Method}) optimizes prompts for neural IR-to-assembly compilation, differing from general APO techniques by uniquely learning from the complete self-debugging trails.
To the best of our knowledge, there are no research studies that specifically investigate the impact of automatic prompt learning on driving LLMs to generate assembly code. Experiments (Section~\ref{sec:Experiment}) show that the proposed method effectively enables LLMs to learn from past practices, enhancing both the correctness and optimization quality of the generated assembly code.

\section{Neural Compilation Task}
\label{sec:NeuralCompilationTask}

\subsection{Problem Formulation}

Let $P_{source}$ denote a program expressed in a source representation, which can be either a high-level programming language $\mathcal{L}_{HL}$ (e.g., C, Python) or an Intermediate Representation (IR) $\mathcal{L}_{IR}$ (e.g., LLVM IR). Let $A_{target}$ be the corresponding program represented in a low-level assembly language $\mathcal{L}_{ASM}$ specific to a target hardware architecture $\mathcal{M}$ (e.g., x86\_64, aarch64).

Formally, the task of \textbf{Neural Compilation} is defined as learning a mapping function $f_{NC}$, parameterized by a Large Language Model (LLM) with parameters $\theta$, which directly translates an input program $P_{source}$ into an output assembly program $A_{target}$: $A_{target} = f_{NC}(P_{source}, \mathcal{M}; \theta), \quad P_{source} \in \{\mathcal{L}_{HL}, \mathcal{L}_{IR}\}, \quad A_{target} \in \mathcal{L}_{ASM}$.

This translation explicitly conditions on the target architecture $\mathcal{M}$, as assembly syntax and semantics are inherently architecture-dependent. A successful \emph{Neural Compilation} system must satisfy two critical objectives simultaneously:

\begin{itemize}
  \item \textbf{Functional Correctness}: The generated assembly program $A_{\text{target}}$ must maintain semantic equivalence to the source program $P_{\text{source}}$. Formally, this means that for any valid input or execution context, the observable behavior and final computational state must match precisely: $\llbracket A_{\text{target}} \rrbracket_{\mathcal{M}} \;\equiv\; \llbracket P_{\text{source}} \rrbracket$, where $\llbracket \cdot \rrbracket$ denotes the semantic interpretation of a program, and $\llbracket \cdot \rrbracket_{\mathcal{M}}$ specifically refers to execution semantics on the target hardware architecture $\mathcal{M}$.

  \item \textbf{Performance Optimization}: Beyond correctness, the generated assembly code should exhibit high performance. Performance metrics include execution speed (runtime), instruction count, code size, memory footprint, and energy consumption. Ideally, the assembly output from the LLM ($A_{\text{target}}$) should be competitive with or outperform code produced by highly optimized traditional compilers, such as those using the \texttt{-O3} flag: $c\bigl(A_{\text{target}}\bigr) \;\le\; c\bigl(A_{\text{compiler(O3)}}\bigr)$, where $c(\cdot)$ denotes a performance cost function applicable to the target architecture.
  
\end{itemize}

Building on this formulation, we introduce a foundational workflow for \emph{Neural Compilation} that systematically addresses both correctness and performance, drawing from common patterns in LLM-based code generation and optimization. The workflow follows the general structure below:

\begin{itemize}

  \item \textbf{Initial Generation}: Given an input program \( P_{\text{source}} \), the LLM generates an initial assembly candidate \( A^{(0)}_{\text{target}} \) conditioned on the target architecture \( \mathcal{M} \). If the initial generation fails to produce a functionally correct program, the process terminates early.
  
  \item \textbf{Iterative Optimization}: Starting from \( A^{(0)}_{\text{target}} \), the model performs up to \( T \) rounds of optimization. In each round \( t = 1, 2, \dots, T \), a new candidate \( A^{(t)}_{\text{target}} \) is generated based on \( A^{(t-1)}_{\text{target}} \), aiming to improve performance.

\end{itemize}

Note that after each generation or optimization step, a self-debugging procedure may be applied to verify and enforce functional correctness.

\subsection{NeuComBack Dataset}
\label{sec:NeuComBackDataset}

\subsubsection{Data Collecting}
Our primary objective is to construct a comprehensive dataset for the LLVM IR to Assembly (ASM) compilation task. To the best of our knowledge, no publicly available datasets specifically focus on this IR-to-ASM translation. Consequently, we adapted existing C-to-ASM benchmarks. We selected \texttt{ExeBench}~\citep{armengol2022exebench}, a widely used collection of C programs, and the Test Suite for Vectorizing Compilers (\texttt{TSVC})~\citep{maleki2011evaluation}, a compiler performance benchmark. The distinct characteristics of these sources naturally lead to a two-tiered dataset:
\begin{itemize}
    \item \textbf{Level 1 (Fundamental Compilation, 200 tasks):}
    This selection from the \texttt{ExeBench} test suite covers a broad variety of real-world C programs and features diverse control-flow patterns. A significant portion of these programs, derived from embedded systems applications, exhibits intensive I/O operations. While extremely complex control flow (e.g., deep recursion, concurrency) is underrepresented in available sources, L1 is designed with a breadth-first focus to assess general-purpose functional correctness.

    \item \textbf{Level 2 (Optimization Potential, 151 tasks):} This level, drawn from the \texttt{TSVC} benchmark suite, features programs characterized by computationally simple execution paths but notable loop intricacy. Such programs provide a strong basis for assessing the optimization capabilities reflected in the generated assembly.
\end{itemize}

We curated two benchmark levels: NeuComBack-L1, comprising 200 \texttt{ExeBench} test cases selected for the longest IR lengths from a cleaned set of 1,618; and NeuComBack-L2, the entire \texttt{TSVC} benchmark (151 cases). While both levels have similar lines of code, Clang AST analysis shows \texttt{TSVC} programs average significantly more variables (26.06 vs. 13.73 for \texttt{ExeBench}). Crucially for LLM-based compilation, \texttt{TSVC}'s nested loop structures (prevalent in L2) present far greater difficulty than ifs or gotos due to complex register allocation and scheduling, making L2 a more demanding benchmark.

\subsubsection{Data Cleaning}
The original \texttt{ExeBench} dataset presented challenges in terms of code quality and complexity. 
Many programs were notably short (as illustrated in Appendix~\ref{app:exebench_code_stats}, Figure~\ref{fig:exebench-stat-appendix}), and the original authors employed \texttt{g++ -fpermissive} for compilation due to numerous deviations from C/C++ standards.
To address this, we focused on the \texttt{ExeBench} test set, performing substantial rewriting to ensure C/C++ standard compliance. A critical step in our data cleaning pipeline was to rigorously verify that \texttt{clang} could successfully compile each modified program into LLVM IR.

Furthermore, to mitigate the issue of programs being overly simplistic or short, we filtered the cleaned \texttt{ExeBench} data. We specifically selected programs that, when compiled by \texttt{clang}, produced the longest LLVM IR sequences. 
This selection process enriches our Level 1 dataset with more substantial and representative examples for the IR-to-ASM task, resulting in a set of 200 programs.

\subsubsection{Evaluation Metrics}
The dataset adopts two primary metrics, designed to reflect the dual goals of \emph{Neural Compilation}: achieving functional correctness and generating high-performance assembly code.
\begin{itemize}

    \item \textbf{Functional Correctness (ACC):} This metric measures the percentage of test cases where the LLM-generated assembly code produces the exact same output as the reference assembly code compiled by \texttt{clang -O0} (or another appropriate baseline for correctness) for a given set of inputs. For the \texttt{TSVC} dataset, we enhanced its original correctness verification methodology. While the existing checks might suffice for traditional compilers, they are less adequate for LLMs, where errors can manifest at any instruction. Our improved validation is more rigorous in detecting such fine-grained discrepancies.
    \item \textbf{Correctness with Superior Performance (ACC+Perf):} This metric quantifies the percentage of programs that are not only functionally correct but also exhibit execution performance superior to that achieved by \texttt{clang -O3}. This directly assesses the LLM's capability to generate optimized assembly.
\end{itemize}

For \texttt{ExeBench}-derived cases, we reuse the dataset's official inputs and checkers, with only lightweight wrappers adapted to accommodate our IR$\rightarrow$ASM setting.
For \texttt{TSVC}, to avoid false positives from degenerate constant initializations, we replace fixed-value array initializations with deterministic pseudorandom values drawn under a fixed seed, providing diverse inputs while preserving reproducibility.

Performance evaluations are conducted exclusively on the Level 2 (\texttt{TSVC}) dataset, as the Level 1 (\texttt{ExeBench}) dataset, though containing diverse real-world programs, was not primarily designed for performance evaluation, making its programs unsuitable for such analysis.
To ensure reliable performance measurements, we execute each program (both LLM-generated and the \texttt{clang -O3} baseline) 11 times.
We then discard the first three (warm-up) and last three (cool-down) runs, retain runs 4–8 (middle five by execution order), and report the median of these five execution times (median computed by their execution time cost) to mitigate system noise and provide a stable performance figure.

\begin{figure}[htbp]
\vspace{-5ex}
\centering
\makebox[\textwidth][c]{
  \includegraphics[width=1.0\textwidth]{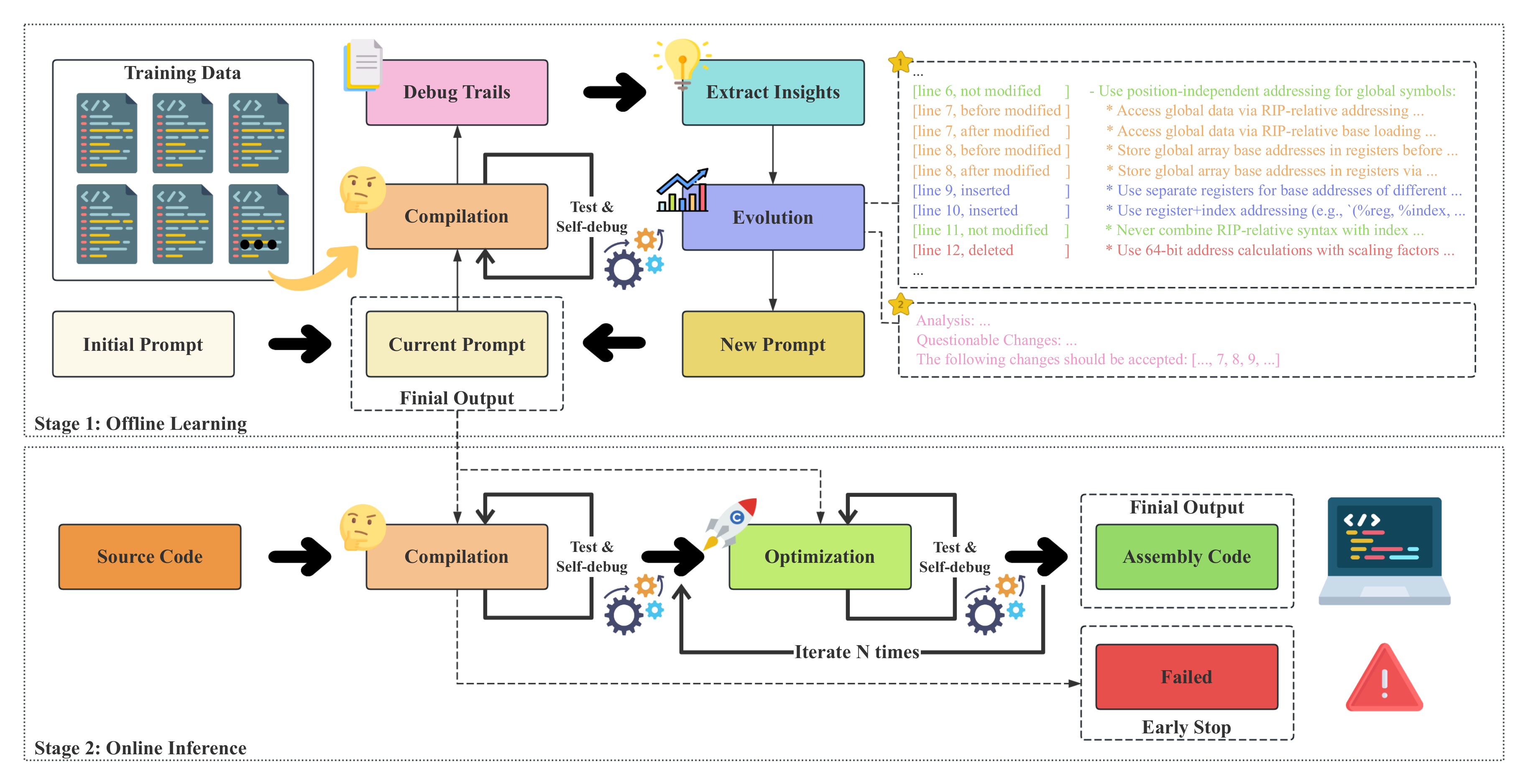}%
}
\vspace{-4ex}
\caption{Pipeline of our automatic prompt learning method on \emph{Neural Compilation}.}
\label{fig:Neural_Compilation}
\vspace{-3ex}
\end{figure}

\section{Method}
\label{sec:Method}

\subsection{Overview}

Our \emph{Neural Compilation} framework leverages a novel automatic prompt-learning mechanism specifically tailored to improve LLM-generated assembly code. 
Distinct from previous prompt evolution methods, our approach's core insight lies in learning from the LLM's complete iterative self-debugging process, which effectively enables the LLM to learn from its past practices in diagnosing and resolving complex errors in assembly code generation. 
Our method comprises two distinct stages, illustrated in Figure~\ref{fig:Neural_Compilation}: offline prompt learning and online inference.
In the offline stage, the model iteratively evolves prompts based on comprehensive analysis and insights derived from past generation trials. 
Subsequently, in the online stage, the model utilizes these refined prompts to generate and iteratively optimize assembly code, progressively improving quality and performance.

\subsection{Self-Evolving Prompt Optimization for Neural Compilation}

\subsubsection{Offline: Prompt Learning}

The offline learning phase focuses on automatically discovering and evolving prompts to effectively guide the LLM toward generating correct and performant assembly code.
This process begins with \textbf{initializing} an empty prompt template.
Following this, \textbf{self-debug trials} are collected by using the LLM to perform a ``compilation'' process and then test the generated assembly translations; errors trigger iterative self-debugging, refining code until correctness or iteration limits are reached.
Subsequently, \textbf{critical insights} are extracted by analyzing the complete self-debug trajectories to identify error patterns and effective strategies via LLM-assisted analysis.
Finally, \textbf{prompt evolution} occurs by refining prompts through the integration of experience and insights (extracted from a batch of self-debug trials), which are then reviewed for clarity and effectiveness.

Concretely, we maintain a single evolving prompt $\pi$. During offline learning, after each mini-batch of compilation tasks we perform one update of $\pi$: we collect self-debugging trajectories from those programs that were initially incorrect but became correct via self-debug, and we ask the same model used for compilation to act as a “prompt optimizer.” A meta-prompt instructs it to analyze the trajectories, extract error patterns and effective fixes, and propose edits to the existing prompt $\pi$ to reflect these new insights. After the LLM suggests revisions, a comparison and confirmation step follows: we prompt the LLM to review the change list and confirm only the necessary and beneficial edits, adding stability to the evolution process. The updated $\pi$ is then used for the next mini-batch.

\subsubsection{Online: Inference}

The online inference stage deploys evolved prompts from offline learning to iteratively generate and optimize assembly code.
First, the \textbf{initial assembly generation} is performed, using evolved prompts to generate initial assembly code from IR. This initial assembly is then tested, and iterative self-debugging is triggered until functional correctness is achieved; if self-debug fails after all attempts, the generation is terminated, and a failure is reported.
Next, \textbf{iterative optimization} is applied to further refine the initial assembly code, with evolved prompts still provided to minimize error introduction. Each optimization iteration includes testing and self-debugging as necessary to ensure continued correctness.
Finally, the process \textbf{outputs} the optimized assembly code, demonstrating competitive or superior performance compared to compiler-generated code.

\section{Experiments}
\label{sec:Experiment}

\subsection{Empirical Evaluation of Existing Large Language Models}
\label{sec:BaselineLLMEval}

\begin{wraptable}{r}{0.4\textwidth}
\vspace{-3ex}
\caption{Baseline performance of advanced LLMs, NeuComBack-L2 (overall 151 cases), x86\_64}
\label{tab:main-baseline-llms}
\centering
\vspace{-1ex}
\resizebox{0.4\textwidth}{!}{
    \begin{tabular}{ccc}
    \toprule
    Model & ACC (\%) & ACC+Perf (\%) \\
    \midrule
    GPT-4o        & 1.99 (3/151)          & 0.66 (1/151)         \\
    O3-Mini       & 21.19 (32/151)         & 5.30 (8/151)          \\
    O1            & 19.87 (30/151)         & 5.30 (8/151)          \\
    DeepSeek-V3   & 14.57 (22/151)         & 3.31 (5/151)          \\
    DeepSeek-R1   & 45.70 (69/151)         & 21.85 (33/151)         \\
    \bottomrule
    \end{tabular}
}
\vspace{-3ex}
\end{wraptable}

To establish a performance baseline for current LLMs on the \emph{Neural Compilation} task, we conducted an empirical study evaluating several state-of-the-art LLMs. The primary objective was to assess their capability in translating programs from an IR to correct and performant x86\_64 assembly code. We utilized NeuComBack-L2 as the main test data for our experiment. For each of the 151 test cases, each LLM was tasked to generate x86\_64 assembly code, involving an initial attempt followed by a maximum of two rounds of self-debugging. We evaluated five models: GPT-4o~\citep{openai2024gpt4o}, O3-Mini~\citep{openai2025O3mini}, O1~\citep{openai2024O1}, DeepSeek-V3~\citep{liu2024deepseek}, and DeepSeek-R1~\citep{guo2025deepseek}. The prompt for baseline assessment is detailed in Appendix~\ref{app:baseline_prompt}.
The results are summarized in Table~\ref{tab:main-baseline-llms}. DeepSeek-R1 demonstrated the strongest baseline, achieving 45.70\% functional correctness and outperforming O3 in 21.85\% of cases. Other reasoning models like O3-Mini and O1 also showed notable capabilities. Among models without specialized, extensive reasoning training, DeepSeek-V3 performed best.

The experimental results lead to several key observations. Firstly, advanced LLMs, particularly those enhanced with reasoning capabilities like DeepSeek-R1, exhibit substantially improved performance in the \emph{Neural Compilation} task of IR-to-assembly translation compared to previous models (e.g., GPT-4o). DeepSeek-R1's ability to correctly generate assembly for nearly half of the benchmark cases, and for a significant portion of those to be more performant than -O3, is a promising development. Secondly, even for the best-performing model, achieving functional correctness remains a significant hurdle, and surpassing traditional compiler optimizations is an even greater challenge. The performance of DeepSeek-V3 suggests that foundational model capabilities are improving. These baseline results highlight both the potential of state-of-the-art LLMs in compiler tasks and the substantial room for improvement necessary for practical, widespread adoption.

\subsection{Evaluation of Self-Evolving Prompt Optimization}
\label{sec:OurMethodEval}

\begin{wraptable}{r}{0.38\textwidth}
    \centering
    \vspace{-7.6ex}
    \caption{Performance of automatic prompt learning vs. baseline on NeuComBack-L1 (test set, 40 samples), x86\_64, DeepSeek-R1.}
    \label{tab:ablation-exebench}
    \vspace{-1ex}
    \resizebox{0.23\textwidth}{!}{
    \begin{tabular}{ccc}
    \toprule
    Method & ACC (\%) \\
    \midrule
    Baseline       & 50.00 (20/40)     \\
    Our            & 80.00 (32/40)     \\
    \bottomrule
    \end{tabular}
    }
    \vspace{-3ex}
\end{wraptable}

Building on this empirical study, we evaluate the impact of our proposed automatic prompt learning method, using the DeepSeek-R1 model due to its superior performance. 
We assess its effectiveness on both NeuComBack-L1  and NeuComBack-L2, targeting the x86\_64 architecture.
The NeuComBack-L1 dataset was divided into 120 training samples, 40 for validation, and 40 for testing, while NeuComBack-L2 comprised 101 training samples, 25 for validation, and 25 for testing. For both datasets, we conducted prompt learning over three epochs with a batch size of 5, additionally introducing 1 self-debugging round per generation for NeuComBack-L1 and 2 rounds for NeuComBack-L2.
From this process, we selected the highest-performing prompt based on validation metrics (referred to as the "Learned Prompt"), which we compare against the baseline prompt detailed in Appendix~\ref{app:baseline_prompt}.

Consequently, for NeuComBack-L1, we evaluate functional correctness (ACC) as its programs, derived from \texttt{ExeBench}, serve as fundamental test cases for basic compilation correctness. For NeuComBack-L2, sourced from \texttt{TSVC} and featuring programs with higher loop complexity and optimization potential, we assess both ACC and ACC+Perf (correctness with superior performance over \texttt{clang -O3}).

\textbf{Results on NeuComBack-L1.}
As shown in Table~\ref{tab:ablation-exebench}, the application of the learned prompt led to substantial improvements: the number of functionally correct solutions increased from 20 (50.00\% ACC) with the baseline prompt to 32 (80.00\% ACC), with a relative 60\% improvement.

\begin{wraptable}{r}{0.55\textwidth}
\vspace{-3ex}
\caption{Performance of automatic prompt learning vs. baseline prompt on NeuComBack-L2 (test set, 25 samples), x86\_64, DeepSeek-R1.}
\vspace{-1ex}
\label{tab:main-learned-iterative-x86}
\centering
\resizebox{0.55\textwidth}{!}{
    \begin{tabular}{ccc}
    \toprule
    Method \& Stage & ACC (\%) & ACC+Perf (\%) \\
    \midrule
    \textbf{Baseline Prompt} & &  \\
    \textit{After Initial Generation}     & 44.00 (11/25)  & 24.00 (6/25)        \\
    \textit{After 2 Rounds Iter. Opt.}    & --             & 28.00 (7/25)        \\
    \midrule
    \textbf{Our Method (Learned Prompt)} & & \\
    \textit{After Initial Generation}     & 64.00 (16/25)  & 40.00 (10/25)       \\
    \textit{After 2 Rounds Iter. Opt.}    & --             & 56.00 (14/25)       \\
    \bottomrule
    \end{tabular}
}

\vspace{-4ex}
\end{wraptable}

\textbf{Results on NeuComBack-L2.}
We conducted two sets of experimental setups:
(a) using a baseline prompt for both initial generation and subsequent optimization, and (b) using the learned prompt (see Appendix~\ref{app:learned_prompt}), performing up to two self-debug rounds after each generation or optimization step."
Table~\ref{tab:main-learned-iterative-x86} details the results on the NeuComBack-L2 benchmark ("--" indicates no change compared to the previous stage).
In the initial generation phase, our method increased correctly solved programs on the NeuComBack-L2 test set from 11 (44.00\% ACC) with the baseline prompt to 16 (64.00\% ACC), achieving a 45\% relative improvement, and solutions outperforming -O3 increased from 6 to 10 (a relative improvement of approximately 67\%).
When evaluating the full \emph{Neural Compilation} workflow, we observe that the advantages of the learned prompt are not merely preserved but substantially amplified through iterative optimization. The baseline prompt produced 7 solutions (28\% ACC+Perf) that outperformed the O3 optimization level, whereas the learned prompt consistently generated 14 such solutions (56\% ACC+Perf), yielding a 100\% relative improvement in high-performance assembly code generation. This significant improvement arises from the learned prompt’s ability to guide the model effectively across the entire compilation process. By steering the model away from common suboptimal decisions and error patterns, the learned prompt enhances the optimizer’s overall effectiveness, leading to consistently superior results.

A noteworthy fact is that, among the 16 programs correctly generated using learned prompts, 14 (87.5\%) were further optimized to surpass \texttt{clang-O3} performance (Detailed speedups are reported in Appendix~\ref{app:detailed_runtime}.).
This remarkably high conversion rate demonstrates that when properly guided through iterative refinement, LLMs can achieve not just functional correctness but genuine optimization mastery in complex assembly-level tasks, a capability that appears substantially underutilized in current approaches. Furthermore, the consistent performance improvements across both datasets validate the robustness and general applicability of our prompt learning methodology, suggesting its potential for broader adoption in low-level code optimization scenarios.

Meanwhile, the consistent positive impact across both datasets also underscores the robustness and general applicability of our learning approach.

\subsection{Ablation}

\begin{wraptable}{r}{0.34\textwidth}
\vspace{-3ex}
\caption{Effectiveness of automatic prompt learning, DeepSeek-R1, NeuComBack-L2 (test set, 25 samples), aarch64}
\label{tab:ablation-aarch64}
\vspace{-1ex}
\centering
\resizebox{0.34\textwidth}{!}{
    \begin{tabular}{ccc}
    \toprule
    Method & ACC (\%) & ACC+Perf (\%) \\
    \midrule
    Baseline       & 36.00 (9/25)         & 8.00 (2/25)          \\
    Our            & 72.00 (18/25)        & 28.00 (7/25)         \\
    \bottomrule
    \end{tabular}
}
\vspace{-2ex}
\end{wraptable}

\textbf{Method Effectiveness on Different Instruction Sets.}
To evaluate adaptability to different hardware, we experimented with the aarch64 instruction set using the DeepSeek-R1 model and our NeuComBack-L2 dataset (101 train, 25 validation, 25 test). The learning process involved 3 epochs with 4 self-debugging rounds per generation. The prompt was chosen the same. As detailed in Table~\ref{tab:ablation-aarch64}, this approach significantly enhanced performance on the test set: functional correctness doubled from 9 to 18 solutions (a 100\% relative improvement), and solutions surpassing -O3 performance increased from 2 to 7 (a 250\% relative improvement), demonstrating effectiveness for aarch64.

\begin{wraptable}{r}{0.58\textwidth}
\vspace{-3ex}
\centering
\caption{Performance on NeuComBack-L1 (x86\_64, DeepSeek-R1) using different prompt strategies, showing test set and overall ACC}
\label{tab:prompt-transferability}
\vspace{-1ex}
\resizebox{0.58\textwidth}{!}{
\begin{tabular}{ccc}
\toprule
Prompt Strategy  & Test Set ACC (\%) & Overall ACC (\%) \\
\midrule
Default Prompt            & 50.00 (20/40)      & 54.50 (109/200)   \\
Learned on NeuComBack-L1  & 80.00 (32/40)      & -                 \\
Learned on NeuComBack-L2  & 67.50 (27/40)      & 74.50 (149/200)   \\
\bottomrule
\end{tabular}
}
\vspace{-2ex}
\end{wraptable}

\textbf{Transferability of Learned Prompts across Different Data Distributions.}
We investigated prompt transferability by applying an x86\_64 prompt learned on NeuComBack-L2 directly to NeuComBack-L1 (120 training, 40 validation, 40 test samples) without further learning. As shown in Table~\ref{tab:prompt-transferability}, the NeuComBack-L2-transferred prompt achieved 67.50\% functional correctness on the NeuComBack-L1 test set, a 35\% relative improvement over the default prompt's 50.00\%. While still lower than the 80.00\% from prompt learned specifically on NeuComBack-L1, it demonstrates positive transferability, suggesting the learned prompts can capture generalizable patterns across program distributions.

\begin{wraptable}{r}{0.58\textwidth}
\vspace{-3ex}
\centering
\caption{Average self-debug rounds for successfully compiled programs by DeepSeek-R1 on test sets, comparing our method with the baseline. Lower is better.}
\label{tab:ablation-self-debug-rounds}
\vspace{-1ex}
\resizebox{0.58\textwidth}{!}{%
\begin{tabular}{ccccc}
\toprule
\multirow{2}{*}{Architecture} & \multirow{2}{*}{Dataset} & \multirow{2}{*}{Max Debug Rounds} & \multicolumn{2}{c}{Avg. Self-Debug Rounds} \\
\cmidrule(l){4-5}
& & & Baseline & Our Method \\ \midrule
\multirow{2}{*}{x86\_64} & NeuComBack-L1 & 1 & 0.90 & 0.28 \\
& NeuComBack-L2 & 2 & 1.09 & 0.25 \\
\midrule
aarch64 & NeuComBack-L2 & 4 & 1.44 & 1.22 \\
\bottomrule
\end{tabular}
}
\vspace{-1ex}
\end{wraptable}

\textbf{Effectiveness in Reducing Self-Debug Rounds.} 
We analyzed the average number of self-debug rounds for tested programs that were eventually solved correctly. This evaluation was performed across different architectures and datasets, comparing our method against the baseline. As shown in Table~\ref{tab:ablation-self-debug-rounds}, our method consistently reduces the average number of self-debug rounds needed.
These results indicate that, in addition to achieving higher final functional correctness (as demonstrated in previous sections), our automatic prompt learning approach enables the LLM to converge on correct assembly code more efficiently, requiring fewer self-correction attempts.

\section{Case Studies}
\label{sec:InDepthAnalysis}

\textbf{Analysis of LLM-generated Code on Performance}
As shown in our experiment (Section~\ref{sec:Experiment}), LLMs demonstrate a promising ability to optimize assembly code, sometimes achieving performance even superior to traditional compilers.
Below, we present examples that illustrate the specific optimization techniques employed by LLM.
For instance, LLMs can enhance performance by reducing the instruction count.
Consider the NeuComBack-L2 function \texttt{s452} as an example, shown in Figure~\ref{fig:self-optimize}. 
While both the original LLVM-generated version and the LLM-optimized version exhibit effective vectorization, the former contained redundant addition operations in its detailed implementation.
These redundancies could be eliminated by pre-calculating constants.
The LLM successfully implemented this optimization, notably reducing two \texttt{paddd} (packed doubleword add) instructions to a single one within the inner loop.
This demonstrates the LLM's capability to iteratively refine and further optimize its own generated code.

Another illustrative example is \texttt{s332}, shown in Figure~\ref{fig:optimize-better-o3}.
In this case, the LLM achieves optimizations superior to those produced by LLVM's \texttt{O3} optimization level.
When tasked with finding the first index of an array element satisfying an inequality condition, LLVM \texttt{O3} employs sequential comparisons within the inner loop, without leveraging vectorization.
In contrast, the LLM leverages vector instructions such as \texttt{cmpps} (compare packed single-precision floating-point values).
This enables the simultaneous comparison of four floating-point numbers within the inner loop.
Furthermore, it employs the \texttt{bsfl} (bit scan forward, least significant) instruction to efficiently extract the index from the resulting comparison mask, thereby achieving functional equivalence with the original C code but with significantly improved performance.

\textbf{Analysis of the Learned Prompts}
The effectiveness of our automatic prompt learning method hinges on the quality and specificity of the evolved prompts. To shed light on what the LLM learns and incorporates into its guiding instructions, this subsection provides a content analysis of the learned prompts.
A complete example of a learned prompt, illustrating these aspects in full, can be found in Appendix~\ref{app:learned_prompt}.

In our iterative prompt engineering process, the LLM was progressively enhanced by integrating a comprehensive set of summarized rules. These rules were meticulously designed to address various dimensions of code generation, aiming to systematically elevate the quality, consistency, and accuracy of the generated outputs. By codifying best practices into structured guidelines, the process not only refined the LLM’s syntactic and semantic understanding but also established a robust framework for maintaining alignment with specific coding standards and conventions. 

\begin{wraptable}{r}{0.7\textwidth}
    \vspace{-4ex}
    \centering
    \caption{Cases of rules learned}
    \vspace{-1ex}
    \resizebox{0.7\textwidth}{!}{
        \begin{tabular}{p{2cm}|p{10cm}}
        \toprule
              \multirow{2}{*}{Formatting} & Placing function code exclusively in .text section \newline
              Positioning .size directives immediately after function bodies\\
              \hline
              \multirow{2}{*}{Syntatic}
               & Using .L prefix with exact spelling for local labels \newline
              Append @PLT suffix to external function calls (e.g., `call dummy@PLT`)
              \\
              \hline
              \multirow{2}{*}{Semantic}
               & Clear return registers (XORL \%eax,\%eax) for void functions \newline Pass stack-based parameters in reverse order with alignment padding \\
               \bottomrule
        \end{tabular}
    }
    \label{tab:prompt-example}
    \vspace{-3ex}
\end{wraptable}

These rules spanned multiple aspects of code generation, as shown in Table~\ref{tab:prompt-example}:
Firstly, formatting rules were established, mandating critical principles like the distinct separation of data and code segments, and the precise IEEE 754 binary representation for floating-point numbers.
Secondly, syntactic rules were introduced to govern structural correctness, exemplified by the specifications for XMM register indexing and conventions.
Thirdly, semantic rules provided guidance and best-practice recommendations regarding the nuanced usage of particular instructions, thereby improving the LLM's understanding of instruction semantics.

\section{Conclusion}

This paper advances the field of Neural Compilation by introducing a dedicated IR-to-Assembly benchmark and a novel self-evolving prompt optimization method, enabling more systematic and rigorous evaluation of LLM-based compilation techniques.
We first introduce NeuComBack, a benchmark specifically designed for IR-to-assembly compilation, and conduct a comprehensive evaluation of state-of-the-art LLMs to establish performance baselines. We further propose a self-evolving prompt optimization method that enables LLMs to iteratively improve their compilation strategies by learning from self-debugging traces. Our experimental results demonstrate significant improvements in both functional correctness (increasing from 44\% to 64\% on x86\_64 and from 36\% to 58\% on aarch64) and optimization performance (with 87.5\% of correct x86\_64 programs surpassing \texttt{clang-O3}). These consistent gains across multiple architectures and benchmark distributions validate the effectiveness of our approach and its potential to advance low-level neural compilation.

While our method achieves promising results, there remains room for improvement in both functional correctness and optimization performance. Future work will focus on further refining these LLM-driven compilation techniques and expanding benchmarks to encompass more complex real-world compilation scenarios.

\section{Acknowledgement}

This work is partially supported by the Strategic Priority Research Program of the Chinese Academy of Sciences (Grants No.XDB0660300, XDB0660301, XDB0660302), NSF of China (Grants No.62525203, 62302483, U22A2028, 6240073476), the Science and Technology Major Special Program of Jiangsu (Grant No. BG2024028), CAS Project for Young Scientists in Basic Research (YSBR-029) and Youth Innovation Promotion Association CAS.


\section*{References}

\bibliography{reference}
\bibliographystyle{reference}


\newpage
\appendix

\section{Original ExeBench C Code (Test Set) Statistics}
\label{app:exebench_code_stats}

\begin{figure}[htbp]
    \centering
    \includegraphics[width=0.8\linewidth]{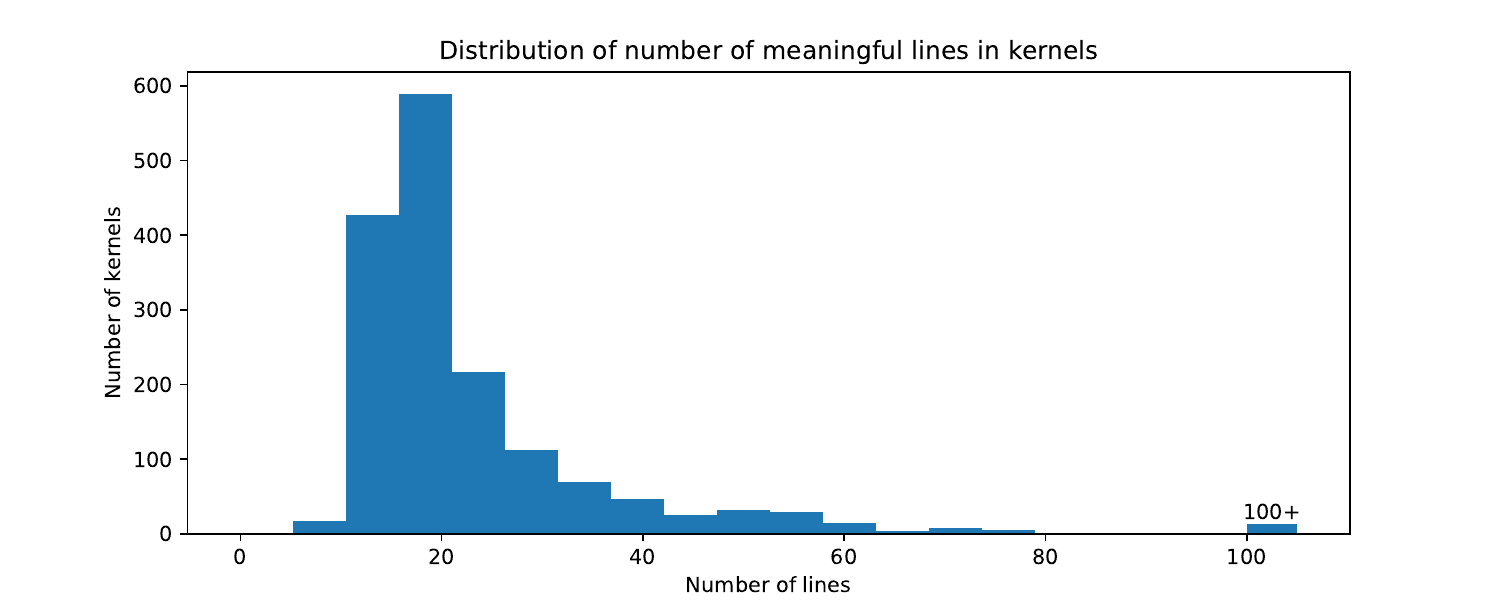}
    \caption{Original \texttt{ExeBench} C code statistics. This figure illustrates the distribution of C code lines in the programs from the original \texttt{ExeBench} dataset (Test Set).}
    \label{fig:exebench-stat-appendix}
\end{figure}

The original \texttt{ExeBench} dataset, prior to our cleaning and filtering process, contained a large number of C programs. As discussed in Section~\ref{sec:NeuComBackDataset} (specifically the Data Cleaning subsubsection), many of these programs were notably short. Figure~\ref{fig:exebench-stat-appendix} provides a visual representation of the C code line count distribution within the original \texttt{ExeBench} dataset (test set).

\section{Baseline Prompt}
\label{app:baseline_prompt}

This appendix shows the baseline prompt utilized for the x86\_64 architecture. This prompt serves as the initial, unoptimized set of instructions given to the Large Language Model for translating LLVM IR to AT\&T format assembly code.

\begin{lstlisting}
```llvm ir
...
```

```assembly
...
```

You are an expert in IR code and assembly code. Please help me translate the given IR code into x86_64 GNU assembly code (using AT&T format).

Note that:
1. You MUST use the following template to give out the complete, final target assembly code, and MUST NOT apply this template to any other part of your response:
```assembly
...(Provide the assembly code here)
```
2. Do not return any information other than the assembly code.
\end{lstlisting}

\section{Example of Learned Prompt}
\label{app:learned_prompt}

This appendix presents an example of a prompt that was evolved using our automatic prompt learning method. Specifically, this prompt is the result of the learning process conducted on the x86\_64 architecture using the NeuComBack-L2 dataset, following the experimental practices described in Section~\ref{sec:OurMethodEval}.

\begin{lstlisting}
```llvm ir
...
```

```assembly
...
```

You are an expert in IR code and assembly code. Please help me translate the given IR code into x86_64 GNU assembly code (using AT&T format).

Note that:
1. You MUST use the following template to give out the complete, final target assembly code, and MUST NOT apply this template to any other part of your response:
```assembly
...(Provide the assembly code here)
```
2. Do not return any information other than the assembly code.
3. Additionally, to guarantee the correctness of the generated assembly code, please ensure that:
    - Correctly structure sections and assembler directives by:
        * Placing function code exclusively in .text section and constant data in .rodata (or other appropriate sections)
        * Maintaining strict section boundaries no intermixing of code and data declarations between sections
        * Positioning .size directives immediately after function bodies within .text section, before any data/comm declarations or section transitions
        * Placing .comm directives exclusively in BSS context (typically .bss section), never within .text or .rodata
        * Using .L prefix with exact spelling for local labels and ensuring case-sensitive consistency in all references
    - Use position-independent addressing for global symbols:
        * Access all global data exclusively via RIP-relative addressing using `leaq symbol(%rip), %reg`
        * Store global array base addresses in registers via LEAQ before indexed accesses, maintaining separate registers for distinct arrays
        * Use register+index addressing (e.g., `(%base_reg, %index, scale)`) with scaling factors matching element size (4 for 32-bit floats/ints)
        * Never combine absolute symbol names with index registers; use base registers initialized via LEAQ for all indexed accesses
    - Maintain proper register usage and stack management:
        * Preserve callee-saved registers (rbx/r12-r15) via push/pop sequences when modified, with dedicated registers for persistent data like array bases and loop counters
        * Use 64-bit registers (rsi/rdi/etc) for loop counters and indices when handling ranges exceeding 32-bit capacity, initializing with XORQ for zeroing
        * Store loop counters in callee-saved registers separate from array pointers, using distinct registers for different control variables
        * Maintain 16-byte stack alignment by calculating adjustments as (pushed_registers*8 + parameters_size + 15) & ~15 before calls. Ensure alignment persists after push/sub operations, especially when storing XMM registers with movaps
        * Transfer final results to return registers (XMM0 for floats, RAX/EAX for integers) BEFORE restoring callee-saved registers in function epilogue
        * Clear return registers (XORL %eax,%eax) for void functions before RET and validate all exit paths set EAX/RAX
        * Explicitly extend 32-bit values to 64-bit via MOVSLQ/CDQE when using 32-bit operations with 64-bit registers
    - Handle function calls and external references correctly:
        * Append @PLT suffix to external function calls (e.g., `call dummy@PLT`)
        * Preserve XMM registers XMM6-XMM15 across calls via save/restore sequences if reused post-call. When storing computed values in these registers, save originals to stack with movaps (aligned) or movups (unaligned), then restore before returning
        * Load parameters into registers via `leaq symbol(%rip)` immediately before call setup to minimize register pressure
        * Pass stack-based parameters in reverse order with alignment padding, recalculating offsets after stack adjustments to ensure correct argument positioning
        * Never assume XMM0-5 retain values across function calls; explicitly preserve results in callee-saved registers if needed
    - Maintain strict data type consistency:
        * Match IR operations with correct instruction types (e.g., use addss for float addition vs. addl for integers)
        * Represent immediate float constants as hexadecimal literals (e.g., 0x3f800000 for 1.0f) instead of separate memory constants when possible
        * Handle residual elements using same data type as vectorized operations (XMM registers for floats)
        * Use MOVSS for 32-bit float transfers between memory/XMM registers instead of integer MOV variants
    - Implement loop semantics accurately:
        * Place loop initializations inside headers when IR indicates per-iteration requirements (PHI-node dependencies)
        * Reinitialize array bases/values in outer loops if inner loops modify global state
        * Maintain separate vectorized (16-element) and scalar residual processing paths
        * Track register/memory modifications through nested loops to prevent cross-iteration dependencies
    - Handle vector operations correctly:
        * Use SIMD registers matching operation width (XMM for 128-bit ops)
        * Verify memory alignment against IR attributes before using aligned accesses (movaps)
        * Default to unaligned accesses (movups) when alignment isn't explicitly guaranteed
        * Process residual elements after vectorized blocks using scalar operations matching original data type
\end{lstlisting}

\section{Detailed Runtime Performance Results}
\label{app:detailed_runtime}

Table~\ref{tab:detailed_runtime} reports per-program speedups of our method (learned-prompt, DeepSeek-R1) relative to \texttt{clang -O3} for the 16 functionally-correct programs referenced in Section~\ref{sec:OurMethodEval} (after initial generation and 2 rounds of iterative optimization). 
Speedup is defined as $\text{Speedup}=\tfrac{\text{Runtime}_{\text{O3}}}{\text{Runtime}_{\text{LLM}}}$, so values $>1$ indicate our code runs faster than \texttt{-O3}. 
Runtimes follow the same protocol as in Section~\ref{sec:Experiment}: each program is executed 11 times and we take the median of the middle 5 runs.

\begin{table}[h]
\centering
\caption{Per-program speedups (LLM over \texttt{clang -O3}) on NeuComBack-L2 (x86\_64). Average speedup across these 16 programs is $1.28\times$; maximum is $3.25\times$ (s331).}
\label{tab:detailed_runtime}
\begin{tabular}{lc}
\toprule
Program & Speedup \\
\midrule
s331   & 3.2485 \\
s1232  & 2.0621 \\
s332   & 1.9954 \\
s112   & 1.1048 \\
s152   & 1.0852 \\
s2251  & 1.0543 \\
s451   & 1.0288 \\
s319   & 1.0278 \\
vdotr  & 1.0277 \\
s3112  & 1.0275 \\
s4121  & 1.0274 \\
s231   & 1.0253 \\
s113   & 1.0247 \\
s422   & 1.0224 \\
s452   & 0.8621 \\
s341   & 0.7924 \\
\bottomrule
\end{tabular}
\end{table}

\section{Illustrative Examples of LLM-Driven Code Optimization}
\label{app:optimization_code_examples}

\begin{figure}[htbp]
    \centering
    \begin{subfigure}[b]{0.4\textwidth}
    \centering
                \includegraphics[width=\linewidth]{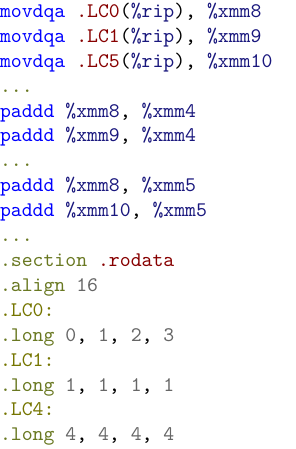}
    \caption{original version}
    \label{fig:self-opt-a}
    \end{subfigure}
    \hspace{0.1\textwidth}
    \begin{subfigure}[b]{0.4\textwidth}
    \vfill
                \includegraphics[width=\linewidth]{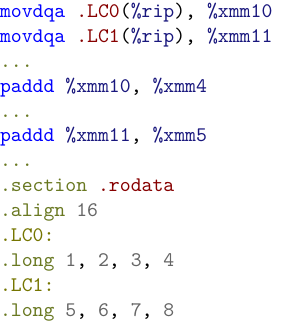}
                \vfill
                \caption{optimized version}
                \label{fig:self-opt-b}
    \end{subfigure}
    \caption{A case of performance self-optimization (\texttt{s452})}
    \label{fig:self-optimize}
\end{figure}

\begin{figure}[htbp]
    \centering
    \begin{subfigure}[b]{0.4\textwidth}
    \centering
                \includegraphics[width=\linewidth]{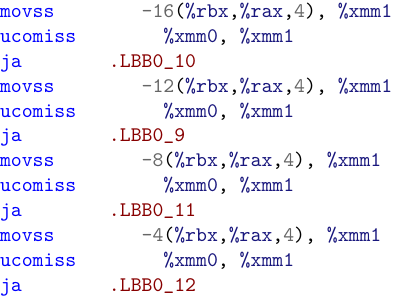}
    \caption{original version}
    \label{fig:self-opt-a}
    \end{subfigure}
    \hspace{0.1\textwidth}
    \begin{subfigure}[b]{0.4\textwidth}
    \vfill
                \includegraphics[width=\linewidth]{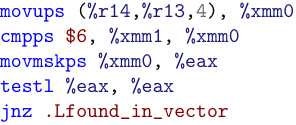}
                \vfill
                \caption{optimized version}
                \label{fig:self-opt-b}
    \end{subfigure}
    \caption{A case of performance better than O3 (\texttt{s332})}
    \label{fig:optimize-better-o3}
\end{figure}

This appendix provides the visual representations (Figures \ref{fig:self-optimize} and \ref{fig:optimize-better-o3}) for the code optimization case studies discussed in Section~\ref{sec:InDepthAnalysis}.
Figure~\ref{fig:self-optimize} showcases the optimization of function \texttt{s452}, where the Large Language Model (LLM) reduced the instruction count.
Figure~\ref{fig:optimize-better-o3} illustrates the optimization of function \texttt{s332}, where the LLM achieved performance superior to LLVM O3 by employing vector instructions. These figures compare the original LLVM-generated assembly code with the LLM-optimized version.

\section{Cost-Controlled Model Comparison}
\label{app:cost_control}

We first estimated the relative inference cost and found that a single experimental run with DeepSeek-R1 is approximately 14 times expensive than with DeepSeek-V3.
Based on this finding, we conducted a cost-controlled evaluation. We allocated DeepSeek-V3 a 14-fold computational budget, allowing it to perform more sampling, to match the cost of a single run of DeepSeek-R1. All other experimental settings remain the same as those used in Section~\ref{sec:OurMethodEval} on NeuComBack-L2.
In our original experiments, DeepSeekR1 solved 11 out of 25 problems (achieving functional correctness). However, under this new cost-controlled setting, DeepSeekV3 successfully solved 17 out of 25 problems.

\begin{table}[h]
\centering
\caption{Cost-controlled comparison on NeuComBack-L2 (test set, 25 cases).}
\label{tab:cost_control_comp}
\begin{tabular}{lc}
\toprule
Setting & Problems Solved (ACC) \\
\midrule
DeepSeek-R1 (single run) & 11/25 \\
DeepSeek-V3 ($14\times$ sampling) & 17/25 \\
\bottomrule
\end{tabular}
\end{table}

\section{Limitations}
While this work advances \emph{Neural Compilation} with a new benchmark and a self-evolving prompt optimization method, we also highlight several limitations.

\paragraph{IR scope.}
Our study focuses on LLVM IR, for it is the most general Intermediate Representations, used by various programming languages, so the techniques are likely to generalize to other IRs, particularly those based on Static Single Assignment (SSA) form; however, generalization is not guaranteed. The design choices and assumptions in our method may not hold equally across all IR designs, especially those with significantly different semantics or control-flow structures.

\paragraph{Benchmark scale.}
\texttt{NeuComBack} is moderate in size and not intended for pre-training or fine-tuning. It is designed primarily for evaluation: each task includes a verifiable test harness for automatic correctness (and, for L2, performance) measurement, and the dataset size keeps the cost/time of running frontier models practical and acceptable.

\paragraph{Benchmark diversity.}
\texttt{NeuComBack-L1} focuses on basic correctness; \texttt{NeuComBack-L2} (TSVC-derived) emphasizes loop-centric, vectorization-friendly kernels that expose clear optimization headroom. This leaves less coverage of other compiler challenges (e.g., pointer-heavy code, irregular control flow, recursion, dynamic memory, interprocedural/code-layout optimizations, register pressure, and memory-hierarchy effects). Consequently, ACC+Perf gains on L2 should not be over-interpreted as universal across all optimization classes.

We will work to address these limitations in future work.


\newpage
\section*{NeurIPS Paper Checklist}

\begin{enumerate}

\item {\bf Claims}
    \item[] Question: Do the main claims made in the abstract and introduction accurately reflect the paper's contributions and scope?
    \item[] Answer: \answerYes{}
    \item[] Justification: The abstract and introduction accurately describe the paper's contributions, including the NeuComBack dataset, the baseline LLM evaluations, the proposed automatic prompt learning method, and the experimental results demonstrating its effectiveness in improving IR-to-assembly compilation. These are detailed in Sections 1 (Introduction), 3.2 (NeuComBack Dataset), 4 (Method).

\item {\bf Limitations}
    \item[] Question: Does the paper discuss the limitations of the work performed by the authors?
    \item[] Answer: \answerNo{}
    \item[] Justification: The paper does not have a dedicated ``Limitations'' section. However, potential limitations that could be discussed include:
    \begin{itemize}
        \item \textbf{Scope of \texttt{NeuComBack} Benchmark:} The current \texttt{NeuComBack} benchmark primarily leveraging \texttt{ExeBench} and \texttt{TSVC} as data sources and currently focuses on the IR-to-assembly translation task. Future work could expand it with more data and additional compilation-related tasks (e.g., different IRs, other optimization stages).
        \item \textbf{Extensibility of the Prompt Learning Method:} The proposed automatic prompt learning method has shown strong results. Nevertheless, it could be further extended by exploring more sophisticated prompt evolution strategies, potentially incorporating techniques like genetic algorithms or other search heuristics to navigate the prompt space more broadly.
        \item \textbf{Computational Cost:} The automatic prompt learning process, involving multiple LLM inference calls for generation, self-debugging, and insight extraction, can be computationally intensive. The associated costs might pose a practical challenge for very large-scale prompt searches or when using highly expensive LLM APIs.
        \item \textbf{Generalizability to Weaker Models:} The current method demonstrates significant effectiveness with recent, advanced LLMs (e.g., DeepSeek-R1). Its efficacy and the transferability of learned insights might be more limited when applied to smaller or less capable models, which would require further investigation.
    \end{itemize}

\item {\bf Theory assumptions and proofs}
    \item[] Question: For each theoretical result, does the paper provide the full set of assumptions and a complete (and correct) proof?
    \item[] Answer: \answerNA{}
    \item[] Justification: The paper presents an empirical study with a new benchmark dataset (NeuComBack) and a novel methodology (automatic prompt learning) for assembly generation. It does not propose new theoretical results that would require formal assumptions and proofs. The problem formulation in Section 3.1 defines the task but does not introduce theorems.

    \item {\bf Experimental result reproducibility}
    \item[] Question: Does the paper fully disclose all the information needed to reproduce the main experimental results of the paper to the extent that it affects the main claims and/or conclusions of the paper (regardless of whether the code and data are provided or not)?
    \item[] Answer: \answerYes{}
    \item[] Justification: The paper describes the dataset creation process (Section 3.2), evaluation metrics (Section 3.2.3), LLMs used (Section 5.1), experimental setup including data splits (Section 5.1, 5.2), prompt learning parameters (epochs, batch size, self-debug rounds, Section 5.1), and architectures tested (x86\_64, aarch64). Examples of baseline and learned prompts are also stated to be in Appendices A and B, which are crucial for understanding the inputs to the LLM.

\item {\bf Open access to data and code}
    \item[] Question: Does the paper provide open access to the data and code, with sufficient instructions to faithfully reproduce the main experimental results, as described in supplemental material?
    \item[] Answer: \answerNo{}
    \item[] Justification: We plan to release our dataset and code upon acceptance; currently no public repository is linked. We will include a link in the final version.

\item {\bf Experimental setting/details}
    \item[] Question: Does the paper specify all the training and test details (e.g., data splits, hyperparameters, how they were chosen, type of optimizer, etc.) necessary to understand the results?
    \item[] Answer: \answerYes{}
    \item[] Justification: Section 5 (Experiment) details the data splits for NeuComBack-L1 and NeuComBack-L2, the LLMs used (specifically DeepSeek-R1 for the core method evaluation), the number of epochs and batch sizes for prompt learning, the number of self-debugging rounds allowed per generation, and how the ``learned prompt'' was selected (best performing on the validation set). This provides a good understanding of the experimental procedure.
    \item[] Guidelines:t
    \begin{itemize}
        \item The answer NA means that the paper does not include experiments.
        \item The experimental setting should be presented in the core of the paper to a level of detail that is necessary to appreciate the results and make sense of them.
        \item The full details can be provided either with the code, in appendix, or as supplemental material.
    \end{itemize}

\item {\bf Experiment statistical significance}
    \item[] Question: Does the paper report error bars suitably and correctly defined or other appropriate information about the statistical significance of the experiments?
    \item[] Answer: \answerNo{}
    \item[] Justification: The paper reports performance metrics as percentages and raw counts (e.g., "28.00\% (7/25)") in tables such as Table 1, 2, 3, 4, and 5. However, it does not include error bars, confidence intervals, or formal statistical significance tests for these results. Conducting multiple runs for each experimental condition to gather data for robust statistical significance analysis was deemed prohibitively expensive. This is primarily due to the high cost of LLM API calls, particularly when employing advanced reasoning models for complex, token-intensive tasks such as IR-to-assembly translation, where each sample can consume a significant number of tokens.

\item {\bf Experiments compute resources}
    \item[] Question: For each experiment, does the paper provide sufficient information on the computer resources (type of compute workers, memory, time of execution) needed to reproduce the experiments?
    \item[] Answer: \answerNo{}
    \item[] Justification: The paper specifies the Large Language Models used (e.g., DeepSeek-R1, GPT-4o). It is clarified that model inference for these LLMs was performed via API calls. Consequently, the specific underlying compute hardware (e.g., GPU type, memory on the provider's side) for the inference step is managed by the API providers.
    
\item {\bf Code of ethics}
    \item[] Question: Does the research conducted in the paper conform, in every respect, with the NeurIPS Code of Ethics \url{https://neurips.cc/public/EthicsGuidelines}?
    \item[] Answer: \answerYes{}
    \item[] Justification: The research focuses on LLM-based code generation for compilers, using publicly available benchmarks (ExeBench, TSVC) to create a new dataset. It does not involve human subjects, sensitive personal data, or applications with immediate high-risk ethical concerns outlined in the NeurIPS Code of Ethics. The goal is to advance compiler technology.

\item {\bf Broader impacts}
    \item[] Question: Does the paper discuss both potential positive societal impacts and negative societal impacts of the work performed?
    \item[] Answer: \answerNo{}
    \item[] Justification: The Introduction (Section 1) of the paper extensively discusses several potential positive societal impacts of Neural Compilation. These include its potential to streamline compiler development for novel architectures, reduce the time and effort needed for creating compilers for emerging Instruction Set Architectures (ISAs), facilitate the discovery of innovative optimization techniques, and transform how programming language researchers and hardware architects explore new designs. However, the paper does not explicitly discuss potential negative societal impacts. Examples of such impacts that could be relevant include the potential for deskilling in the compiler engineering workforce if traditional roles are significantly altered, the risk of misuse if the technology were adapted to generate obfuscated or malicious code (though this is not the paper's focus), or the environmental impact related to the energy consumption of training and deploying large LLMs for these compilation tasks.
    
\item {\bf Safeguards}
    \item[] Question: Does the paper describe safeguards that have been put in place for responsible release of data or models that have a high risk for misuse (e.g., pretrained language models, image generators, or scraped datasets)?
    \item[] Answer: \answerNA{}
    \item[] Justification: The paper utilizes existing Large Language Models (e.g., DeepSeek-R1, GPT-4o) and introduces a benchmark dataset (NeuComBack) derived from public coding benchmarks. It does not release new foundational pretrained language models or image generators itself, nor does it use scraped datasets that would pose a high risk for misuse in terms of sensitive content or PII. The primary artifacts are the methodology and the benchmark.

\item {\bf Licenses for existing assets}
    \item[] Question: Are the creators or original owners of assets (e.g., code, data, models), used in the paper, properly credited and are the license and terms of use explicitly mentioned and properly respected?
    \item[] Answer: \answerYes{}
    \item[] Justification: The paper properly credits the creators of the datasets used: \texttt{ExeBench} \citep{armengol2022exebench} and \texttt{TSVC} \citep{maleki2011evaluation} are cited in Section 3.2.1. According to public sources, both datasets are released under open-source licenses, and their use in this paper complies with the respective licensing terms.

\item {\bf New assets}
    \item[] Question: Are new assets introduced in the paper well documented and is the documentation provided alongside the assets?
    \item[] Answer: \answerYes{}
    \item[] Justification: The paper introduces the NeuComBack dataset, and its construction, sources, levels, and metrics are documented in Section 3.2. The proposed automatic prompt learning method is described in Section 4. The paper itself serves as the primary documentation for these new assets. If these assets were to be released, further practical documentation (e.g., README files) would typically accompany them.

\item {\bf Crowdsourcing and research with human subjects}
    \item[] Question: For crowdsourcing experiments and research with human subjects, does the paper include the full text of instructions given to participants and screenshots, if applicable, as well as details about compensation (if any)? 
    \item[] Answer: \answerNA{}
    \item[] Justification: The research described in the paper does not involve crowdsourcing experiments or research with human subjects.

\item {\bf Institutional review board (IRB) approvals or equivalent for research with human subjects}
    \item[] Question: Does the paper describe potential risks incurred by study participants, whether such risks were disclosed to the subjects, and whether Institutional Review Board (IRB) approvals (or an equivalent approval/review based on the requirements of your country or institution) were obtained?
    \item[] Answer: \answerNA{}
    \item[] Justification: The research does not involve human subjects and therefore did not require IRB approval or an equivalent review.

\item {\bf Declaration of LLM usage}
    \item[] Question: Does the paper describe the usage of LLMs if it is an important, original, or non-standard component of the core methods in this research? Note that if the LLM is used only for writing, editing, or formatting purposes and does not impact the core methodology, scientific rigorousness, or originality of the research, declaration is not required.
    \item[] Answer: \answerYes{}
    \item[] Justification: The use of Large Language Models (LLMs) such as DeepSeek-R1 is a fundamental and core component of this research. The entire paper is about ``Neural Compilation,'' defined as using LLMs to generate assembly code, and proposes an ``automatic prompt learning method'' specifically to enhance LLM performance in this task. This is detailed throughout the paper, particularly in Sections 1 (Introduction), 3 (Neural Compilation Task), 4 (Method), and 5 (Experiment).

\end{enumerate}


\end{document}